\documentclass[runningheads]{llncs}
\usepackage{graphicx}
\usepackage{comment}
\usepackage{amsmath,amssymb} 
\usepackage{color}
\usepackage[percent]{overpic}
\usepackage{hyperref}

\usepackage{bm}

\usepackage{array}
\newcolumntype{L}[1]{>{\raggedright\let\newline\\\arraybackslash\hspace{0pt}}m{#1}}
\newcolumntype{C}[1]{>{\centering\let\newline\\\arraybackslash\hspace{0pt}}m{#1}}
\newcolumntype{R}[1]{>{\raggedleft\let\newline\\\arraybackslash\hspace{0pt}}m{#1}}

\begin{document}
\pagestyle{headings}
\mainmatter
\def\ECCVSubNumber{2985}  
\title{Neural Hair Rendering} 

\titlerunning{Neural Hair Rendering}
%
\author{Menglei Chai \and
Jian Ren \and
Sergey Tulyakov}
\authorrunning{M. Chai et al.}
%
\institute{Snap Inc.}
\maketitle

\begin{abstract}
In this paper, we propose a generic neural-based hair rendering pipeline that can synthesize photo-realistic images from virtual 3D hair models. Unlike existing supervised translation methods that require model-level similarity to preserve consistent structure representation for both real images and fake renderings, our method adopts an unsupervised solution to work on arbitrary hair models. The key component of our method is a shared latent space to encode appearance-invariant structure information of both domains, which generates realistic renderings conditioned by extra appearance inputs. This is achieved by domain-specific pre-disentangled structure representation, partially shared domain encoder layers and a structure discriminator. We also propose a simple yet effective temporal conditioning method to enforce consistency for video sequence generation. We demonstrate the superiority of our method by testing it on a large number of portraits and comparing it with alternative baselines and state-of-the-art unsupervised image translation methods.
\end{abstract}

\section{Introduction}
\label{sec:introduction}
Hair is a critical component of human subjects. Rendering virtual 3D hair models into realistic images has been long studied in computer graphics, due to the extremely complicated geometry and material of human hair. Traditional graphical rendering pipelines try to simulate every aspect of natural hair appearance, including surface shading, light scattering, semi-transparent occlusions, and soft shadowing. This is usually achieved by leveraging physics-based shading models of hair fibers, global illumination rendering algorithms, and artistically designed material parameters. Given the extreme complexity of the geometry and associated lighting effects, such a direct approximation of physical hair appearance requires a highly detailed 3D model, carefully tuned material parameters, and a huge amount of rendering computation. However, for interactive application scenarios that require efficient feedback and user-friendly interactions, such as games and photo editing softwares, it is often too expensive and unaffordable.

With the recent advances in generative adversarial networks, it becomes natural to formulate hair rendering as a special case of the conditional image generation problem, with the hair structure controlled by the 3D model, while realistic appearance synthesized by neural networks. In the context of image-to-image translation, one of the major challenges is how to bridge both the source and target domains for proper translation. Most existing hair generation methods fall into the supervised category, which demands enough training image pairs to provide direct supervision. For example, sketch-based hair generation methods \cite{lee2019maskgan,jo2019sc,qiu2019two} construct training pairs by synthesizing user sketches from real images. While several such methods are introduced, rendering 3D hair models with the help of neural networks do not receive similar treatment. The existing work on this topic \cite{wei2018real} requires real and fake domains considerably overlap, such that the common structure is present in both domains. This is achieved at the cost of a complicated strand-level high-quality model, which allows for extracting edge and orientation maps that serve as the common representations of hair structures between real photos and fake models. However, preparing such a high-quality hair model is itself expensive and non-trivial even for a professional artist, which significantly restricts the application scope of this method.

In this paper, we propose a generic neural-network-based hair rendering pipeline that provides efficient and realistic rendering of a generic low-quality 3D hair model borrowing the material features extracted from an arbitrary reference hair image. Instead of using a complicated strand-level model to match real-world hairs like \cite{wei2018real}, we allow users to use any type of hair model requiring only the isotropic structure of hair strands be properly represented. Particularly, we adopt sparse polygon strip meshes which are much more widely used in interactive applications \cite{ward2007survey}. Given the dramatic difference between such a coarse geometry and real hair, we are not able to design common structure representations at the model level. Therefore, supervised image translation methods will be infeasible due to the lack of paired data.

To bridge the domains of real hair images and low-quality virtual hair models in an unsupervised manner, we propose to construct a shared latent space between both real and fake domains, which encodes a common structural representation from distinct inputs of both domains and renders the realistic hair image from this latent space with the appearance conditioned by an extra reference. This is achieved by 1) different domain structure encodings used as the network inputs, to pre-disentangle geometric structure and chromatic appearance for both real hair images and 3D models; 2) a UNIT \cite{liu2017unsupervised}-like architecture adopted to enable common latent space by partially sharing encoder weights between two auto-encoder branches that are trained with in-domain supervision; 3) a structure discriminator introduced to further match the distribution of the encoded structure features; 4) supervised reconstruction enforced on both branches to guarantee all necessary structure information is kept in the shared feature space. In addition, to enable temporally-smooth animation rendering, we introduce a simple yet effective temporal condition method with single image training data only, utilizing the exact hair model motion fields. We demonstrate the effectiveness of the pipeline and each key component by extensively testing on a large amount of diverse human portraits and various hair models. We also compare our method with general unsupervised image translation methods, and show that due to the limited sampling ability on the synthetic hair domain, all existing methods fail to produce convincing results.

\section{Related Work}
\label{sec:related_work}
\paragraph{Image-to-image translation} aims at converting images from one domain to another while keeping the structure of the source image unchanged. The literature contains various methods performing this task in different settings. Paired image-to-image translation methods \cite{isola2017image,wang2018high} operate when pairs of images in both domains are available. For example, semantic labels to scenes \cite{wang2018high,park2019semantic,chen2017photographic}, edges to objects \cite{sangkloy2017scribbler}, and image super-resolution \cite{ledig2017photo,johnson2016perceptual}. However, paired data are not always available in many tasks. Unsupervised image-to-image translation tackles a setting in which paired data is not available, while sampling from two domains is possible \cite{liu2019few,taigman2016unsupervised,zhu2017unpaired,dundar2018domain,shrivastava2017learning,liu2017unsupervised,huang2018multimodal}. Clearly, unpaired image-to-image translation is an ill-posed problem for there are numerous ways an image can be transformed to a different domain. Hence, recently proposed methods introduce constraints to limit the number of possible transformations. Some studies enforce certain domain properties \cite{bousmalis2017unsupervised,shrivastava2017learning}, while other concurrent works apply cycle-consistency to transform images between different domains \cite{yi2017dualgan,zhu2017unpaired,kim2017learning}. Our work differs from existing studies that we focus on a specific challenging problem, which is the realistic hair generation, where we want to translate manually designed hair models from the domain of rendered images to the domain of real hair. For the purpose of controllable hair generation, we leverage rendered hair structure and arbitrary hair appearance to synthesize diverse realistic hairstyles. The further difference in our work compared to the image-to-image translation papers is unbalanced data. The domain of images containing real hair is far more diverse than that of rendered hair, making it even more challenging for classical image-to-image translation works to address the problem.

\paragraph{Neural style transfer} is related to image-to-image translation in a way that image style is changed while content is maintained \cite{chen2016fast,gatys2016image,huang2017arbitrary,li2016precomputed,li2017diversified,li2017demystifying,ulyanov2016texture,hertzmann2001image}. Style in this case is represented by unique style of an artist \cite{gatys2016image,ulyanov2016texture} or is copied from an example image provided by the user. Our work follows the research idea from example-guide style transfer that hairstyle is obtained from reference real image. However, instead of changing the style of a whole image, our aim is to keep the appearance of the human face and background unchanged, while having full control over the hair region. Therefore, instead of following exiting works that inject style features into image generation networks directly \cite{huang2017arbitrary,park2019semantic}, we propose a new architecture that combines only hair appearance features and latent features that encodes image content and adapted hair structure for image generation. This way we can achieve the goal that only the style of the hair region is manipulated according to the provided exemplar image. 

\paragraph{Domain Adaptation} addresses the domain-shift problem that widely exists between the source and target domains \cite{saenko2010adapting}. Various feature-based methods have been proposed to tackle the problem \cite{kulis2011you,gong2012geodesic,gopalan2011domain,fernando2013unsupervised,tzeng2014deep}. Recent works on adversarial learning for the embedded feature alignment between source and target domains achieve better results than previous studies \cite{ganin2014unsupervised,ganin2016domain,liu2016coupled,tsai2018learning,hoffman2017cycada,tzeng2017adversarial}. Efforts using domain adaptation for both classification and pixel-level prediction tasks have gained significantly progress \cite{bousmalis2017unsupervised,chen2017no,tsai2018learning}. In this work, we follow the challenging setting of unsupervised domain adaptation that there is no corresponding annotation between source and target domains.
We aim at learning an embedding space that only contains hair structure information for both rendered and real domains. Considering the domain gap, instead of using original images as input, we use rendered and real structure map as inputs to the encoders, which contain both domain-specific layers and shared layers, to obtain latent features. The adaptation is achieved by adversarial training and image reconstruction.

\paragraph{Hair Modeling, Rendering, and Generation} share a similar goal with our paper, which is synthesizing photo-realistic hair images. With 3D hair models manually created \cite{ward2007survey,yuksel2009hair}, captured \cite{paris2008hair,herrera2012lighting,luo2013structure,hu2014robust,zhang2017data}, or reconstructed from images \cite{chai2012single,chai2013dynamic,hu2015single,chai2015high,chai2016autohair,zhou2018hairnet}, traditional graphical hair rendering methods focus on improving rendering quality and performance by either more accurately modeling the special hair material and lighting behaviours \cite{marschner2003light,moon2006simulating,deon2011energy,yan2015physically}, or approximating certain aspects of rendering pipeline to reduce the computation complexity \cite{zinke2008dual,moon2008efficient,sadeghi2010artist,ren2010interactive,xu2011interactive}. However, the extremely huge computation cost for realistic hair rendering usually prohibits them to be directly applied in real-time applications. Utilizing the latest advances in GANs, recent works \cite{lee2019maskgan,jo2019sc,qiu2019two,olszewski2020intuitive,tan2020michigan} achieved impressive progress on conditioned hair image generation as supervised image-to-image translation. A GAN-based hair rendering method \cite{wei2018real} proposes to perform conditioned 3D hair rendering by starting with a common structure representation and progressively enforce various conditions. However, it requires the hair model to be able to generate consistent representation (strand orientation map) with real images, which is challenging for low-quality mesh-based models, and cannot achieve temporally smooth results.

\begin{figure}[t]
\begin{center}
\centering
\includegraphics[width=0.75\linewidth]{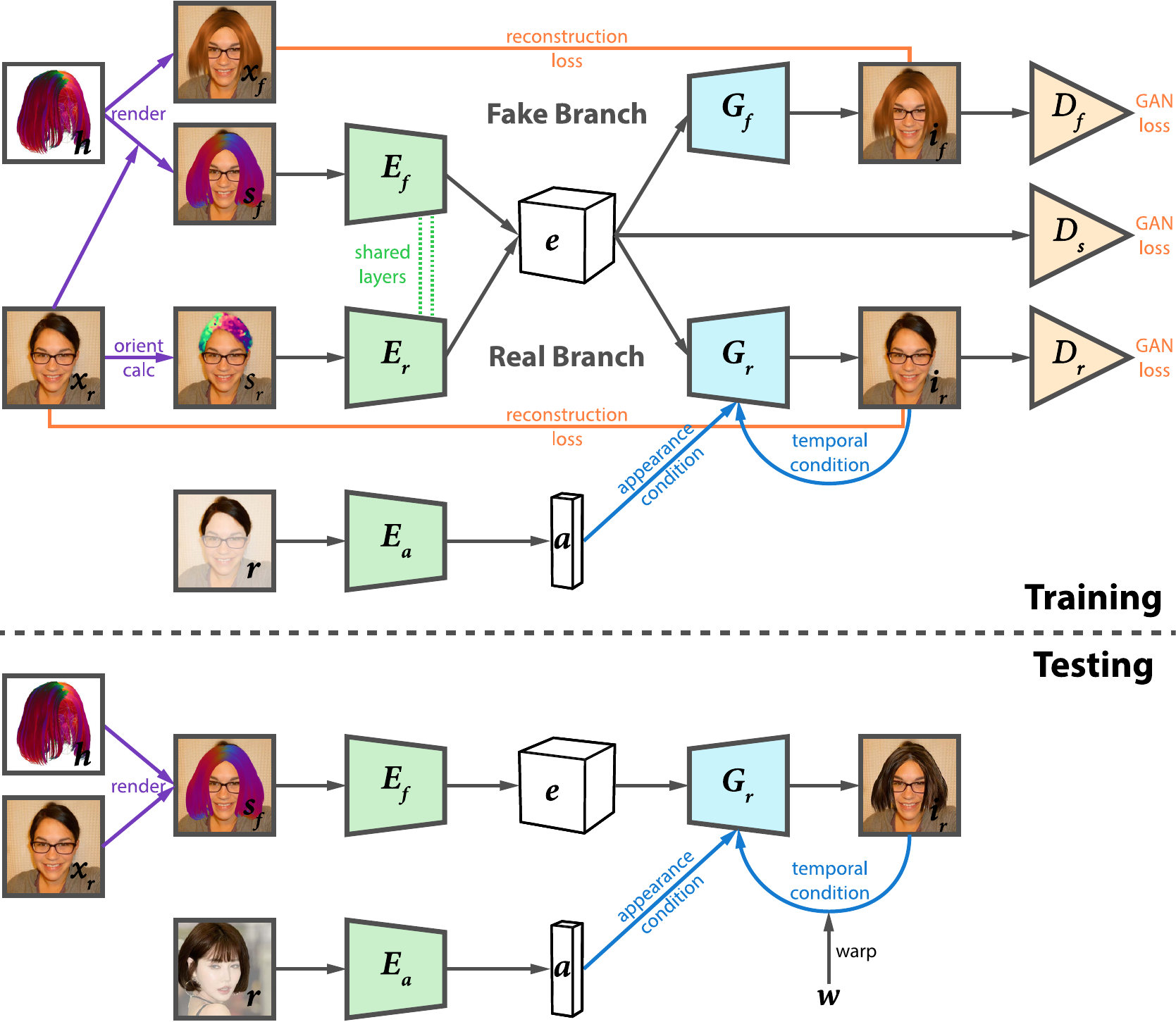}
\caption{\textbf{The overall pipeline of our neural hair rendering framework.} We use two branches to encode hair structure features, one for the real domain and the other for the fake domain. A domain discriminator is applied to the outputs from both encoders, to achieve domain invariant features. We also use two decoders to reconstruct images for two domains. The decoder in the real domain is different from the one in the fake domain, for it is conditioned on a reference image. Additionally, to generate consistent videos, we apply a temporal condition on the real branch. During inference, we use the encoder in the fake branch to get hair structure features from a 3D hair model and use the generator in the real branch to synthesized an appearance conditioned image.}
\label{fig:pipeline}
\end{center}
\end{figure}

\section{Approach}
\label{sec:approach}

Let $\bm{h}$ be the target 3D hair model, with camera parameters $\bm{c}$ and hair material parameters $\bm{m}$, we formulate the \textit{traditional graphic rendering pipeline} as $\text{R}_{t}(\bm{h}, \bm{m}, \bm{c})$. Likewise, our \textit{neural network-based rendering pipeline} is defined as $\text{R}_n(\bm{h}, \bm{r}, \bm{c})$, with a low-quality hair model $\bm{h}$ and material features extracted from an arbitrary reference hair image $\bm{r}$.

\subsection{Overview of Network Architecture}
\label{sec:method_network_architecture}
The overall system pipeline is shown in Fig.\ref{fig:pipeline}, which consists of two parallel branches for both domains of real photo (i.e., \textit{real}) and synthetic renderings (i.e., \textit{fake}), respectively. 

On the encoding side, the \textit{structure adaptation subnetwork}, which includes a real encoder $E_r$ and a fake encoder $E_f$, achieves cross-domain structure embedding $\bm{e}$. Similar to UNIT\cite{liu2017unsupervised}, we share the weights of the last few ResNet layers in $E_r$ and $E_f$ to extract consistent structural representation from two domains. In addition, a structure discriminator $D_s$ is introduced to match the high-level feature distributions between two domains to enforce the shared latent space further to be domain invariant.

On the decoding side, the \textit{appearance rendering subnetwork}, consisting of $G_r$ and $G_f$ for the real and fake domain respectively, is attached after the shared latent space $\bm{e}$ to reconstruct the images in the corresponding domain.
Each decoder owns its exclusive domain discriminator $D_r$ and $D_f$ to ensure the reconstruction matches the domain distribution, besides the reconstruction losses. The hair appearance is conditioned in an asymmetric way that $G_r$ accepts the extra condition of material features extracted from a reference image $\bm{r}$ by using material encoder $E_m$, while the unconditional decoder $G_f$ is asked to memorize the appearance, which is made on purpose for training data generation (Sec.\ref{sec:data_preparation}).
 
At the \textit{training stage}, all these networks are jointly trained using two sets of image pairs $(\bm{s}, \bm{x})$ for both real and fake domains, where $\bm{s}$ represents a domain-specific structure representation of the corresponding hair image $\bm{x}$ in this domain. Both real and fake branches try to reconstruct the image $G(E(\bm{x}))$ from its paired structure image $\bm{s}$ independently through their own encoder-decoder networks, while the shared structural features are enforced to match each other consistently by the structure discriminator $D_s$. We set the appearance reference $\bm{r} = \bm{x}$ in the real branch to fully reconstruct $\bm{x}$ in a paired manner. 

At the \textit{inference stage}, only the fake branch encoder $E_f$ and the real branch decoder $G_r$ are activated. $G_r$ generates the final realistic rendering using structural features encoded by $E_f$ on the hair model. The final rendering equation $\text{R}_n$ can be formulated as:
\begin{equation}
\text{R}_n(\bm{h},\bm{r},\bm{c})=G_r(E_f(\text{S}_f(\bm{h},\bm{c})),E_m(\bm{r})),
\end{equation}
where the function $\text{S}_f(\bm{h},\bm{c})$ renders the structure encoded image $\bm{s}_f$ of the model $\bm{h}$ in camera setting $\bm{c}$.

\subsection{Structure Adaptation}
\label{sec:structure_adaptation}
The goal of the structure adaptation subnetwork, formed by the encoding parts of both branches, is to encode cross-domain structural features to support final rendering.
Since the inputs to both encoders are manually disentangled structure representation (Sec.\ref{sec:data_preparation}), the encoded features $E(\bm{s})$ only contain structural information of the target hair. Moreover, as the appearance information is either conditioned by extra decoder input in a way that non-spatial-varying structural information is leaked (the real branch) or simple enough to be memorized by the decoder (the fake branch) (Sec.\ref{sec:appearance_rendering}), the encoded features should also include all the structural information necessary to reconstruct $\bm{x}$.

$E_r$ and $E_f$ share a similar network structure: five downsampling convolution layers followed by six ResBlks. The last two ResBlks are weight-sharing to enforce the shared latent space. $D_s$ follows PatchGAN\cite{isola2017image} to distinguish between the latent feature maps from both domains:
\begin{equation}
\mathcal{L}_{D_s}=\mathbb{E}_{\bm{s}_r}[\log(D_s(E_r(\bm{s}_r)))]+\mathbb{E}_{\bm{s}_f}[\log(1-D_s(E_f(\bm{s}_f)))].
\end{equation}

\subsection{Appearance Rendering}
\label{sec:appearance_rendering}

The hair appearance rendering subnetwork decodes the shared cross-domain hair features into the real domain images.
The decoders $G_r$ and $G_f$ have different network structures and do not share weights since the neural hair rendering is a unidirectional translation that aims to map the rendered 3D model in the fake domain to real images in the real domain.
Therefore, $G_f$ is required to make sure the latent features $\bm{e}$ encode all necessary information from the input 3D model, instead of learning to render various appearance. On the other hand, $G_r$ is designed in a way to accept arbitrary inputs for realistic image generation.


Specifically, the unconditional decoder $G_f$ starts with two ResBlks, and then five consecutive upsampling transposed convolutional layers followed by one final convolutional layer. $G_r$ adopts a similar structure as $G_f$, with each transposed convolutional layer replaced with a SPADE\cite{park2019semantic} ResBlk to use appearance feature maps $\bm{a}_{\bm{r},\bm{s}_r}$ at different scales to condition the generation. Assuming the binary hair mask of the reference and the target images are $\bm{m}_{\bm{r}}$ and $\bm{m}_{\bm{s}}$, the appearance encoder $E_m$ extracts the appearance feature vector on $\bm{r}\times\bm{m}_{\bm{r}}$ with five downsampling convolutional layers and an average pooling. This feature vector $E_m(\bm{r})$ is then used to construct the feature map $\bm{a}_{\bm{r},\bm{s}_r}$ by duplicating it spatially in the target hair mask $\bm{m}_{\bm{s}}$ as follows:
\begin{equation}
\bm{a}_{\bm{r},\bm{s}_r}(p)=\begin{cases} 
E_m(\bm{r}),  & \mbox{if }\bm{m}_{\bm{s}_r}(p)=1, \\
0, & \mbox{if }\bm{m}_{\bm{s}_r}(p)=0.
\end{cases}
\end{equation}
To make sure the reconstructed real image $G_r(E_r(\bm{s}_r),\bm{a}_{\bm{r},\bm{s}_r})$ and the reconstructed fake image $G_f(E_f(\bm{s}_f))$ belong to their respective distributions, we apply two domain specific discriminator $D_r$  and $D_f$ for the real and fake domain respectively. The adversarial losses write as:
\begin{equation}
\mathcal{L}_{D_r}=\mathbb{E}_{\bm{x}_r}[\log(D_r(\bm{x}_r))]+\mathbb{E}_{\bm{s}_r,\bm{r}}[\log(1-D_r(G_r(E_r(\bm{s}_r),\bm{a}_{\bm{r},\bm{s}_r})))],
\end{equation}
\begin{equation}
\mathcal{L}_{D_f}=\mathbb{E}_{\bm{x}_f}[\log(D_f(\bm{x}_f))]+\mathbb{E}_{\bm{s}_f}[\log(1-D_f(G_f(E_f(\bm{s}_f))))].
\end{equation}
We also adopt perceptual losses to measure high-level feature distance utilizing the paired data:
\begin{equation}
\begin{split} 
\mathcal{L}_p=\sum_{l=0}^L\ &\|\bm{\Psi}_l(G_r(E_r(\bm{s}_r),\bm{a}_{\bm{r},\bm{s}_r}))-\bm{\Psi}_l(\bm{x}_r)\|_1\\+&\|\bm{\Psi}_l(G_f(E_f(\bm{s}_f)))-\bm{\Psi}_l(\bm{x}_f)\|_1,
\end{split}
\end{equation}
where $\bm{\Psi}_l(\bm{i})$ computes the activation feature map of input image $\bm{i}$ at the $l$th selected layer of VGG-19\cite{simonyan2015very} pre-trained on ImageNet\cite{russakovsky2015imagenet}.

Finally, we have the overall training objective as:
\begin{equation}
\label{eq:objective}
\min_{E,G}\max_{D}\ (\lambda_s\mathcal{L}_{D_s}+\lambda_g(\mathcal{L}_{D_r}+\mathcal{L}_{D_f})+\lambda_p\mathcal{L}_p).
\end{equation}

\subsection{Temporal Conditioning}
\label{sec:temporal_conditioning}
The aforementioned rendering network is able to generate plausible single-frame results. However, despite the hair structure is controlled by smoothly-varying inputs of $\bm{s}_f$ with the appearance conditioned by a fixed feature map $\bm{a}_{\bm{r},\bm{s}_r}$, the spatially-varying appearance details are still generated in a somewhat arbitrary manner which tends to flicker in time (Fig.\ref{fig:video_results}). Fortunately, with the availability of the 3D model, we can calculate the exact hair motion flow $\bm{w}^t$ for each pair of frames $t-1$ and $t$, which can be used to warp image $\bm{i}$ from $t-1$ to $t$ as $\text{W}(\bm{i},\bm{w}^t)$. We utilize this dense correspondences to enforce temporal smoothness.

Let $\bm{I}=\{\bm{i}^0,\bm{i}^1,\ldots,\bm{i}^T\}$ be the generated result sequence, we achieve this temporal conditioning by simply using the warped result of the previous frame $\text{W}(\bm{i}^{t-1},\bm{w}^t)$ as an additional condition, stacked with the appearance feature map $\bm{a}_{\bm{r},\bm{s}_r}$, to the real branch decoder $G_r$ when generating the current frame $\bm{i}^t$.

We achieve temporally consistent by changing the real branch decoder only with temporally finetuning. During temporal training, we fix all other networks and use the same objective as Eq.\ref{eq:objective}, but randomly ($50\%$ of chance) concatenate $\bm{x}_r$ into the condition inputs to the SPADE ResBlks of $G_r^t$. The generation pipeline of the real branch now becomes $G_r^t(E_r(\bm{s}_r),\bm{a}_{\bm{r},\bm{s}_r},\bm{x}_r)$, so that the network learns to preserve the consistency if the previous frame is inputted as the temporal condition, or generate randomly from scratch if the condition is zero.

Finally, we have the rendering equation for sequential generation:
\begin{equation}
\begin{split}
\bm{i}^t=\text{R}_n(\bm{h},\bm{r},\bm{c}^t)&=\begin{cases} 
G_r(E_f(\bm{s}_f^t),\bm{a}_{\bm{r},\bm{s}_f^t}),&\mbox{if }t=0,\\
G_r^t(E_f(\bm{s}_f^t),\bm{a}_{\bm{r},\bm{s}_f^t},\text{W}(\bm{i}^{t-1},\bm{w}^t)).&\mbox{if }t>0,
\end{cases}\\
\bm{s}_f^t&=\text{S}_f(\bm{h},\bm{c}^t).
\end{split}
\end{equation}

\section{Experiments}
\label{sec:experiments}
\begin{figure}[t]
\begin{center}
\centering
\includegraphics[width=0.8\linewidth]{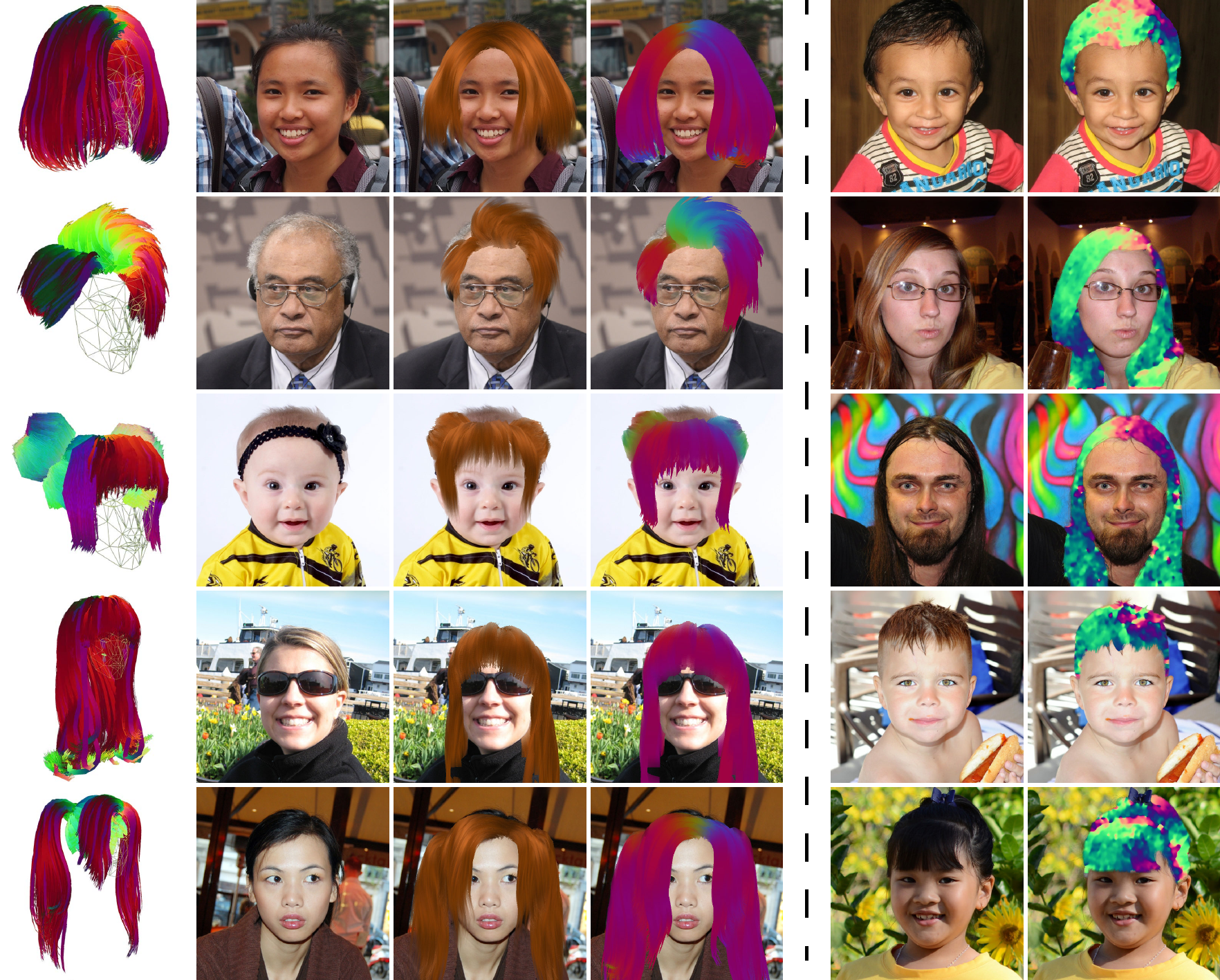}
\setlength{\tabcolsep}{0\linewidth}
\begin{tabular}{C{0.136\linewidth}C{0.136\linewidth}C{0.136\linewidth}C{0.136\linewidth}C{0.034\linewidth}C{0.136\linewidth}C{0.136\linewidth}}
3D Hair&Input&Fake Hair&Structure&&Real Hair&Structure\\ 
\end{tabular}
\begin{tabular}{C{0.544\linewidth}C{0.034\linewidth}C{0.272\linewidth}}
Fake Domain&&Real Domain\\ 
\end{tabular}
\caption{\textbf{Training data preparation.} For the fake domain (left), we use hair model and input image to generate fake rendering and model structure map. For the real domain (b), we generate image structure map for each image.}
\label{fig:training_data}
\end{center}
\end{figure}

\subsection{Data Preparation}
\label{sec:data_preparation}

To train the proposed framework, we generate a dataset that includes image pairs $(\bm{s}, \bm{x})$ for both real and fake domains. In each domain, $\bm{s}\rightarrow \bm{x}$ indicates the mapping from structure to image, where $\bm{s}$ encodes only the structure information, and $\bm{x}$ is the corresponded image that conforms to the structure condition.  

\textbf{Real Domain.} We adopt the widely used FFHQ\cite{karras2019style} portrait dataset to generate the training pairs for the real branch, given it contains diverse hairstyles on shapes and appearances.
To prepare real data pairs, we use original portrait photos from FFHQ as $\bm{x}_r$, and generate $\bm{s}_r$ to encode only structure information from hair. However, obtaining $\bm{s}_r$ is a non-trivial process since hair image also contains material information, besides structural knowledge. To fully disentangle structure and material, and construct a universal structural representation $\bm{s}$ of all real hair, we apply a dense pixel-level orientation map in the hair region, which is formulated as $\bm{s}_r = \text{S}_r(\bm{x}_r)$,  calculated with oriented filter kernels \cite{paris2008hair}. Thus, we can obtain $\bm{s}_r$ that only consists of local hair strand flow structures. Example generated pairs are presented in Fig.\ref{fig:training_data}b.

For the purpose of training and validation, we randomly select $65, 000$ images from FFHQ as training, and use the remaining $5, 000$ images for testing. For each image $\bm{x}_r$, we perform hair segmentation using off-the-shelf model \cite{chai2016autohair}, and calculate $\bm{s}_r$ for the hair region.

\textbf{Fake Domain.} There are multiple ways to model and render virtual hair models. From coarse to fine, typical virtual hair models range from a single rigid shape, coarse polygon strips representing detached hair wisps, to a large number of thin hair fibers that mimic real-world hair behaviors. Due to various granularity of the geometry, the structural representation is hardly shared with each other or real hair images. In our experiments, all the hair models we used are polygon strips based considering this type of hair model is widely adopted in real-time scenarios for it is efficient to render and flexible to be animated. To generate $\bm{s}_f$ for a given hair model $\bm{h}$ and specified camera parameters $\bm{c}$, we use smoothly varying color gradient as texture to render $\bm{h}$ into a color image that embeds the structure information of the hair geometry, such that $\bm{s}_f = \text{S}_f(\bm{h}, \bm{c})$. As for $\bm{x}_f$, we use traditional graphic rendering pipeline to render $\bm{h}$ with a uniform appearance color and simple diffuse shading, so that the final synthetic renderings have a consistent appearance that can be easily disentangled without any extra condition, and keep all necessary structural information to verify the effectiveness of the encoding step. Example pairs are shown in Fig.\ref{fig:training_data}a.

For the 3D hair used for fake data pairs, we create five models (leftmost column in Fig.\ref{fig:training_data}). The first four models are used for training, and the last one is used to evaluate the generalization capability of the network, for the network has never seen it. All these models consist of $10$ to $50$ polygon strips, which is sparse enough for real-time applications. We use the same training set from the real domain to form training pairs. Each image is overlaid by one of the four 3D hair models according to the head position and pose. Then the image with the fake hair model is used to generate $\bm{x}_f$ through rendering the hair model with simple diffuse shading, and $\bm{s}_f$ by exporting color textures that encode surface tangent of the mesh. We strictly use the same shading parameters, including lighting and color, to enforce a uniform appearance of hair that can be easily disentangled by the networks.

\begin{figure}
\begin{center}
\centering
\includegraphics[width=\linewidth]{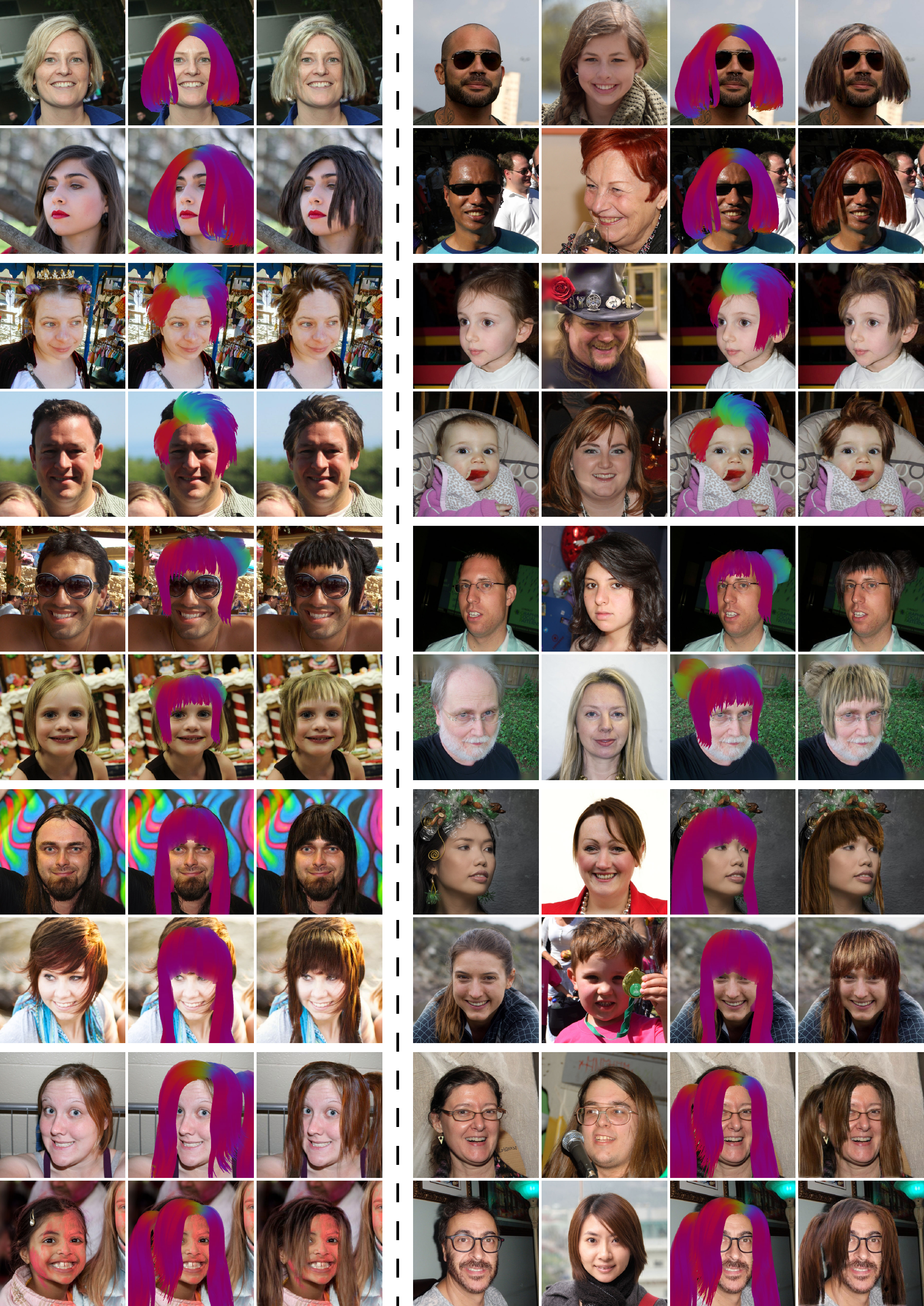}
\setlength{\tabcolsep}{0\linewidth}
\begin{tabular}{C{0.138\linewidth}C{0.138\linewidth}C{0.138\linewidth}C{0.034\linewidth}C{0.138\linewidth}C{0.138\linewidth}C{0.138\linewidth}C{0.138\linewidth}}
Input\&Ref&Structure&Result&&Input&Ref&Structure&Result\\ 
\end{tabular}
\caption{\textbf{Results} for the hair models used in this study (2 rows per model). We visualize examples where the input and the reference image are the same (left), and the input and the reference are different images (right). In the former case the method copies appearance from another image.
}
\label{fig:rendering_results}
\end{center}
\end{figure}

\subsection{Implementation Details}
\label{sec:implementation_details}
We apply a two-stage learning strategy. During the first stage, all networks are trained jointly following Eq.\ref{eq:objective} for the single-image renderer $\text{R}_n$. After that, we temporally fine-tune the decoder $G_r$ of the real branch, to achieve temporally-smooth renderer $\text{R}_n^t$, by introducing the additional temporal condition as detailed in Sec.\ref{sec:temporal_conditioning}. To make the networks of both stages consistent, we keep the same condition input dimensions, including appearance and temporal, but set the temporal condition to zero during the first stage. During the second stage, we set it to zero with $50\%$ of chance. 
The network architecture discussed in Sec.\ref{sec:approach} is implemented using PyTorch.  We adopt Adam solver with a learning rate set to $0.0001$ for the first stage, and $0.00001$ for the fine-tuning stage. The training resolution of all images is $512\times512$, with the mini-batch size set to $4$. For the loss functions, weights $\lambda_p$, $\lambda_s$, and $\lambda_g$ are set to $10$, $1$, and $1$, respectively.
All experiments are conducted on a workstation with $4$ Nvidia Tesla P100 GPUs. During test time, rendering a single frame takes less than $1$ second, with structure encoding less than $200$ms and final generation less than $400$ms.

\subsection{Qualitative Results}
We present visual hair rendering results from two settings in Fig.\ref{fig:rendering_results}. The left three columns in Fig.\ref{fig:rendering_results} show that the reference image $\bm{r}$ is the same as $\bm{x}_r$. By applying a hair model, we can modify human hair shape but keep the original hair appearance and orientation. The right four columns show that the reference image is different from $\bm{x}_r$, therefore, both structure and appearance of hair from $\bm{x}_r$ can be changed at the same time to render the hair with a new style.
We also demonstrate our video results in Fig.\ref{fig:video_results} (please click the image to watch video results online), where we adopt 3D face tracking \cite{cao2018stabilized} to guide the rigid position of the hair model, and physics-based hair simulation method \cite{chai2014reduced} to generate secondary hair motion.
These flexible applications demonstrate that our method can be easily applied to modify hair and generate novel high-quality hair images.


\subsection{Comparison Results}
\label{sec:comparisons}
To the best of our knowledge, there is no previous work that tackles the problem of neural hair rendering; thus, a direct comparison is not feasible. However, in light of our methods aim to bridge two different domains without ground-truth image pairs, which is related to unsupervised image translation, we compare our network with state-of-the-art unpaired image translation studies.
It is important to stress that although our hair rendering translation falls into the range of image translation problems, there exist fundamental differences compared to the generic unpaired image translation formulations for the following two reasons. 

First and foremost, compared with translation between two domains, such as painting styles, or seasons/times of the day, which have roughly the same amount of images for two domains and enough representative training images can be sampled to provide nearly-uniform domain coverage, our real/fake domains have dramatically different sizes--it is easy to collect a huge amount of real human portrait photos with diverse hairstyles to form the real domain. Unfortunately, for the fake domain, it is impossible to reach the same variety since it would require manually designing \textit{every} possible hair shape and appearance to describe the distribution of the whole domain of rendered fake hair. Therefore, we focus on a realistic assumption that only a limited set of such models are available for training and testing, such that we use four 3D models for training and one for testing, which is far from being able to produce variety in the fake domain. 

Second, as a deterministic process, hair rendering should be conditioned strictly on both geometric shape and chromatic appearance, which can be hardly achieved with unconditioned image translation frameworks.

With those differences bearing in mind, we show the comparison between our method and three unpaired image translation studies, including CycleGAN \cite{zhu2017unpaired}, DRIT \cite{lee2018diverse}, and UNIT \cite{liu2017unsupervised}. For the training of these methods, we use the same sets of images,  $\bm{x}_r$ and $\bm{x}_f$, for both real and fake domains, and the default hyperparameters reported by the original papers. Additionally, we compare with the images generated by the traditional graphic rendering pipeline. We denote the method as \textbf{Graphic Renderer}. Finally, we report two ablation studies to evaluate the soundness of the network and the importance of each step: 1) we first remove the structural discriminator (termed as \textbf{w/o SD}); 2) we then additionally remove the shared latent space (termed as \textbf{w/o SL and SD}).

\begin{table}[t]
\centering
\caption{\textbf{Quantitative comparison results}. We compare our method against commonly adopted image-to-image translation frameworks, reporting Fréchet Inception Distance (FID, lower the better), Intersection over Union (IoU, higher the better) and pixel accuracy (Accuracy, higher the better). Additionally we report ablation studies by first removing the structural discriminator (w/o SD) followed by removing both the structural discriminator and the shared latent space (w/o SL and SD). }
\begin{tabular}{C{0.25\linewidth}||C{0.17\linewidth}|C{0.17\linewidth}|C{0.18\linewidth}}
\hline
Method & FID $\downarrow$ & IoU(\%) $\uparrow$ & Accuracy(\%) $\uparrow$\\
\hline
Graphic Renderer & 98.62 & 55.77 & 86.17\\
\hline\hline
CycleGAN \cite{zhu2017unpaired} & 107.11 & 46.46 & 84.06 \\
\hline
UNIT \cite{liu2017unsupervised} & 116.79 & 30.89 &84.27 \\
\hline
DRIT \cite{lee2018diverse} & 174.39 & 30.69 &65.80 \\
\hline \hline
w/o SL and SD & 94.25 & 80.10 &93.89 \\
\hline
w/o SD & 77.09 & 86.60 & 96.35\\
\hline\hline
Ours & \textbf{57.08} & \textbf{86.74}  & \textbf{96.45}\\
\hline
\end{tabular}
\label{tab:ablation}
\end{table}

\subsubsection{Quantitative comparison.}
For quantitative evaluation, we adopt FID (Fréchet Inception Distance) \cite{heusel2017gans} to measure the distribution distance between two domains. Moreover, inspired by the evaluation protocol from existing work \cite{chen2017photographic,wang2018high}, we apply a pre-trained hair segmentation model \cite{svanera2016figaro} on the generated images to get the hair mask, and compare it with the ground truth. Intuitively, the segmentation model should predict the hair mask that similar to the ground-truth for the realistic synthesized images. To measure the segmentation accuracy, we use both Intersection-over-Union (IoU) and pixel accuracy (Accuracy).

The quantitative results are reported in Tab.\ref{tab:ablation}. Our method significantly outperforms the state-of-the-art unpaired image translation works and graphic rendering approach by a large margin for all three evaluation metrics. The low FID score proves our method can generate high-fidelity hair images that contain similar hair appearance distribution as images from the real domain. The high IoU and Accuracy demonstrate the ability of the network to minimize the structure gap between real and fake domains so that the synthesized images can follow the manually designed structure. Furthermore, the ablation analysis in Tab.\ref{tab:ablation} shows both shared encoder layers and the structural discriminator are essential parts of the network, for the shared encoder layers help the network to find a common latent space that embeds hair structural knowledge, while the structural discriminator forces the hair structure features to be domain invariant.

\begin{figure}[t]
\begin{center}
\centering
\includegraphics[width=0.95\linewidth]{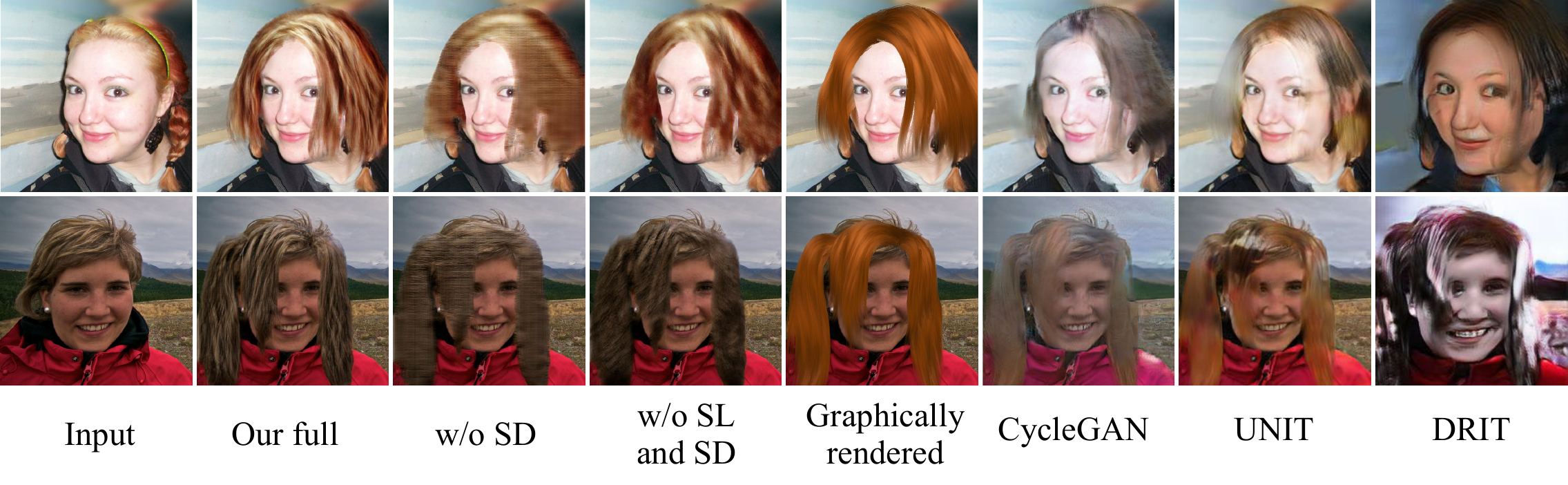}
\caption{\textbf{Visual comparisons.} We show selected visual comparisons against commonly adopted image-to-image translation methods as well as visualize ablation results. Our method synthesizes more realistic hair images compared to other approaches.}
\label{fig:comparisons}
\end{center}
\end{figure}

\subsubsection{Qualitative comparison.}
The qualitative comparison of different methods is shown in Fig.\ref{fig:comparisons}. It can be easily seen that our generated images have much higher quality than the synthesized images created by other state-of-the-art unpaired image translation methods, for they have clearer hair mask, follow hair appearance from reference images, maintain the structure from hair models, and look like natural hair. Compared with the ablation methods (Fig.\ref{fig:comparisons}c and d), our full method (Fig.\ref{fig:comparisons}b) can follow the appearance from reference images (Fig.\ref{fig:comparisons}a) by generating hair with similar orientation.

\begin{figure}[t]
\begin{center}
\centering
\href{https://mlchai.com/files/neural_hair_rendering_video.mp4}{
\includegraphics[width=0.95\linewidth]{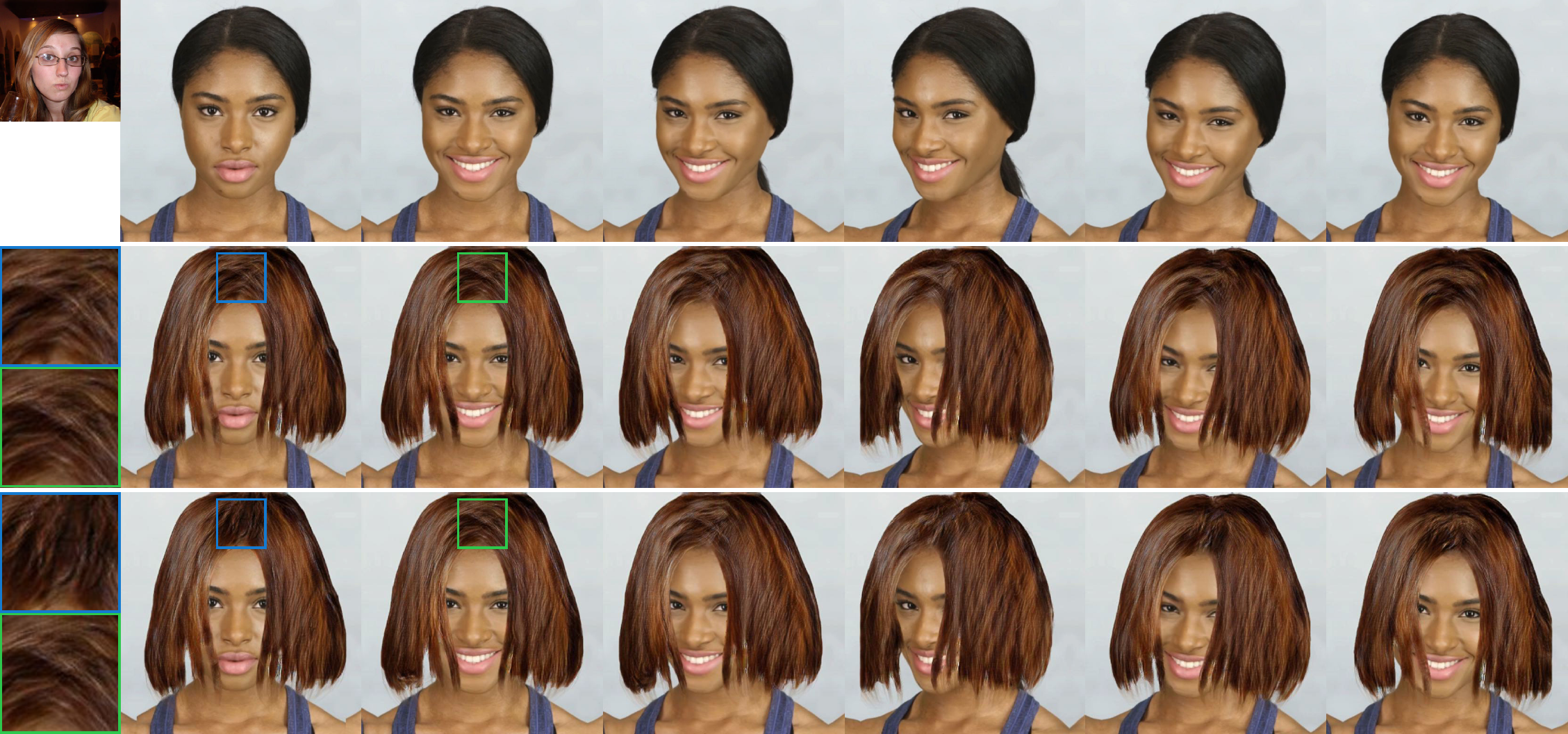}}
\caption{\textbf{Video results and comparisons.} Top row: the first image is the appearance reference image and others are continuous input frames; middle row: generated hair images with temporal conditioning; bottom row: generated hair images without temporal conditioning. We show two zoom-in hair regions for each result. By applying temporal conditioning, our model synthesizes hair images with consistent appearance, while not using temporal conditioning leads to hair appearance flickering as indicated by blue and green boxes. \textbf{\textit{Click the image to play the video results and comparisons.}}}
\label{fig:video_results}
\end{center}
\end{figure}

We also show the importance of temporal conditioning (Sec.\ref{sec:temporal_conditioning}) in Fig.\ref{fig:video_results}. The temporal conditioning helps us generate consistent and smooth video results, for hair appearance and orientation are similar between continuous frames. Without temporal conditioning, the hair texture could be different between frames, as indicated by blue and green boxes, which may result in flickering for the synthesized video. Please refer to the supplementary video for more examples.



\section{Conclusions}
\label{sec:conclusions}
We propose a neural-based rendering pipeline for general virtual 3D hair models. The key idea of our method is that instead of enforcing model-level representation consistency to enable supervised paired training, we relax the strict requirements on the model and adopt an unsupervised image translation framework. To bridge the gap between real and fake domains, we construct a shared latent space to encode a common structure feature space for both domains, even if their inputs are dramatically different. In this way, we can encode a virtual hair model into such a structure feature, and switch it into the real generator to produce realistic rendering. The conditional real generator not only allows flexible appearance conditioning but can also be used to introduce temporal conditioning to generate smooth sequential results.

Our method has several limitations. First, the current method does not change the input. A smaller fake hair won't be able to fully occlude the original one in the input image. It is possible to do face inpainting to remove the excessive hair regions to fix this issue. Second, when the lighting/material of the appearance reference is dramatically different from the input, the result may look unnatural. Better reference selection would help to make the results better. Third, the current method simply blends the generated hair onto the input, which causes blending artifacts in some results especially when the background is complicated. A simple solution is to train a supervised boundary refinement network to achieve better blending quality.

\clearpage
%
%
\bibliographystyle{splncs04}
\bibliography{egbib}
\end{document}